\documentclass[runningheads]{llncs}
\usepackage{graphicx}
\usepackage{float}
\usepackage{dirtytalk}
\usepackage{amsmath}
\usepackage{amsmath,amssymb}

\usepackage{comment}
\usepackage{xspace}
\usepackage{caption}
\usepackage{subcaption}

\begin{document}
\title{Nonnegative Matrix Factorization to understand Spatio-Temporal Traffic Pattern Variations during COVID-19: A Case Study \thanks{This research was supported by the National Research Foundation of Korea (Grant No. 2020R1A2C1012196), and in part by the School of Computer Science and Engineering, Ministry of Education, Kyungpook National University, South Korea, through the BK21 Four Project, AI-Driven Convergence Software Education Research Program, under Grant 4199990214394.}}
\titlerunning{NMF to understand Traffic Pattern Variations during COVID-19}
\author
{
Anandkumar Balasubramaniam\inst{1}\and
Thirunavukarasu Balasubramaniam\inst{2}\and
Rathinaraja Jeyaraj\inst{1}\and 
Anand Paul\inst{1}\and 
Richi Nayak\inst{2}
}
\authorrunning{Anandkumar Balasubramaniam et al.}

\institute{School of Computer Science \& Engineering, Kyungpook National University, Daegu, South Korea. \\ \email{\{bsanandkumar@gmail.com, jrathinaraja@gmail.com, paul.editor@gmail.com\}} \and
School of Computer Science and Centre for Data Science,
Queensland University of Technology,
Brisbane, Australia.\\ \email{\{thirunavukarasu.balas@qut.edu.au, r.nayak@qut.edu.au\}}\\
}

\maketitle              % typeset the header of the contribution
\begin{abstract}
Due to the rapid developments in Intelligent Transportation System (ITS) and increasing trend in the number of vehicles on road, abundant of road traffic data is generated and available. Understanding spatio-temporal traffic patterns from this data is crucial and has been effectively helping in traffic plannings, road constructions, etc. However, understanding traffic patterns during COVID-19 pandemic is quite challenging and important as there is a huge difference in-terms of people's and vehicle's travel behavioural patterns. In this paper, a case study is conducted to understand the variations in spatio-temporal traffic patterns during COVID-19. We apply nonnegative matrix factorization (NMF) to elicit patterns. The NMF model outputs are analysed based on the spatio-temporal pattern behaviours observed during the year 2019 and 2020, which is before pandemic and during pandemic situations respectively, in Great Britain. The outputs of the analysed spatio-temporal traffic pattern variation behaviours will be useful in the fields of traffic management in Intelligent Transportation System and management in various stages of pandemic or unavoidable scenarios in-relation to road traffic.  

\keywords{Traffic Pattern  \and NMF \and Pattern Mining \and Spatio-Temporal Analysis \and COVID-19.}
\end{abstract}
\section{Introduction}
Vehicular traffic pattern analysis is an important topic in developing/improving traffic management system, especially in ITS \cite{in1}. Due to the latest developments and technological advancements in ITS, the availability of the spatio-temporal traffic data is abundant \cite{in2}, \cite{in2a}. However, the traffic patterns are heterogeneous in such a way that various road segments have distinct time-varying traffic patterns \cite{in3}. Addressing this problem of heterogeneity by analysing various features such as location, time, etc., is much needed to get useful traffic pattern insights from this abundance of spatio-temporal traffic data. Also, there are various factors such as congestion, accidents, natural calamities, weather conditions, etc., that affect these vehicular traffic patterns in different ways. 

\par Having said that, the year 2020 seemed to have a lot of uncertainties in people's travel behaviour due to the rapid increase in COVID-19 cases \cite{in4}, \cite{in8}. Due to which, people around the world have faced unimaginable disturbances and struggles that had adversely affected the usual travel patterns of people and vehicles due to the additional enforcement of various stages of curfew. In this paper, vehicular traffic data recorded before and during the pandemic in Great Britain is analyzed; and discussed how the spatio-temporal patterns are getting varied due to the COVID-19 pandemic \cite{in5} in 2020 in Great Britain. %As the spread of COVID-19 is unpredictably faster, the initial cases of COVID-19 in Great Britain is found during the month of January, 2020. From the beginning of the COVID-19 entry in Great Britain, there are 1 in 9 people have been infected according to the analysis \cite{in5}.

\par To understand the spatio-temporal traffic pattern variations due to COVID-19, in this paper, we use NMF \cite{in6}. We apply NMF on traffic data of Grate Britain generated during 2019 and 2020 to have a meaningful insights of spatio-temporal traffic patterns \cite{in7} before and during COVID-19 pandemic i.e. in the year 2019 and 2020 respectively. The case-study provided in this paper helps to identify pattern variations and to understand the impacts of travel or traffic patterns due to COVID-19. While the simple vehicle counts will provide knowledge on how much traffic pattern has changed, it cannot say what patterns have changed. This paper specifically focuses on learning what and how patterns have changed during COVID-19.

The rest of this paper is organized as follows: Section \ref{section2} contains related works, Section \ref{section3} includes the background on NMF, and discuss the traffic pattern elicitation process and the importance of understanding the traffic pattern variations. Section \ref{section4} contains the case study that is conducted on the 2019 and 2020 traffic dataset of Great Britain, which includes dataset descriptions, evaluation measures to run the NMF model followed by spatio-temporal traffic pattern analysis results of 2019 and 2020. Section \ref{section5} concludes the paper and briefly introduces future directions.

\section{Related Work}
Pattern mining is a subset of data mining, which helps to discover or analyse various latent features or insights from high dimensional data. It is commonly applied on vehicular traffic data to analyse the spatio-temporal vehicular traffic patterns \cite{rw1,rw2}. For example, in \cite{rw1}, authors dealt with the problem of identifying  vehicular traffic patterns and congestion problems occurring in densely populated cities. The authors in \cite{rw1} came up with the optimal traffic solution using social media data records. Historical data about social media posts and tweets are taken into consideration and social media knowledge related to the traffic is analysed to predict the traffic pattern at a given location. However, this might not be suitable during pandemic situations like COVID-19 as the mobility pattern of the people changes drastically. In another work \cite{rw2}, authors dealt with the traffic forecasting problem by implementing Fourier analysis with wavelet denoising technique along with two layer Fast Fourier Transform to recognize and formulate the periodic pattern. However, the approach is not capable to elicit spatial patterns.

Similarly, many works have focused on learning traffic patterns to perform different downstream tasks during COVID-19. For instance authors in \cite{rw4} dealt with traffic congestion prevention by learning traffic trends using linear regression. Authors in \cite{rw5} utilized the crowd-sourced traffic congestion data to determine mobility changes in Manhattan, NYC during COVID-19. Time-series decomposition method is used to quantify the changes in various traffic categories. Crowd-sourced traffic data is utilized to inform human mobility changes to plan for the future pandemic responses. The usage of crowd-sourced traffic data might not be reliable due to authenticity of data. Authors in paper \cite{rw6} proposed morphology-based vehicle detection method to detect traffic density in multiple cities and compare traffic data with mobility pattern changes due to COVID-19. Morphology-based vehicle detection technique requires the usage of utilizing image-based dataset, which might not be feasible every-time due to the unavailability of image data. In \cite{rw7}, by using high-resolution remote sensing imagery, traffic density reduction pattern across the various parts of the world during COVID-19 is analysed. Though the results are helpful, the association between temporal and spatial patterns is not learned in their work. 

Recently, factorization-based techniques have been used to learn the associations between spatial and temporal patterns. For instance, authors in \cite{tksnam} used NMF and Nonnegative Tensor Factorization to learn the spatio-temporal dynamics of topics in twitter during COVID-19. In this paper, we investigate the usage of NMF to learn spatio-temporal patterns from traffic data and understand the pattern variations during COVID-19.

\par In this paper, the challenges of combining or comparing the temporal and spatial patterns are addressed in such a way that both the temporal and spatial patterns of the recorded traffic data in Great Britain during the year 2019 and 2020 are analysed and compared along with the utilization of NMF model to understand the spatio-temporal pattern variations before and during COVID-19 pandemic. 
\label{section2}

\begin{comment}
pattern mining, nmf, pattern mining + nmf during COVID-19 for traffic patterns
there is no comparison so we are comparing by applying NMF 
how to apply and perform analysis
In \cite{1}, authors proposed the method for identifying congestion areas and exploring the associated distribution patterns form taxi trajectory data. Clustering methods are employed to detect the congestion-prone areas and congestion clusters interactions are used to visualize the congestion intensity.
\end{comment}

\section{Nonnegative Matrix Factorization for Traffic Pattern Elicitation}
\label{section3}
\subsection{Nonnegative Matrix Factorization}
NMF is one of the popular dimensionality reduction techniques which uses non-negativity constraint to generate non-negative lower-dimensional representations of a matrix, that further lead to the increase in the quality of matrix factorization. Suppose $\boldsymbol{\mathrm{X}} \in {\mathbb{R}} ^{(N \times M)}$ and $\boldsymbol{\mathrm{Y}} \in {\mathbb{R}} ^{(A \times B)}$ are two matrices, they can be factorized into two lower-dimensional factor matrices $\mathrm{U} \in {\mathbb{R}} ^{(N \times r)}$ and $\mathrm{V} \in {\mathbb{R}} ^{(M \times r)}$ \& $\mathrm{P} \in {\mathbb{R}} ^{(A \times k)}$ and $\mathrm{Q} \in {\mathbb{R}} ^{(B \times k)}$, respectively. The factorization of matrix can be defined as, 

\begin{equation}
\label{equation1}
\boldsymbol{\mathrm{X}} \approx \mathrm{U}\mathrm{V}^{T} \quad s.t, \mathrm{U}\geq 0 \quad \textrm{and} \quad \mathrm{V}\geq 0,
\end{equation}

\begin{equation}
\label{equation2}
\boldsymbol{\mathrm{Y}} \approx \mathrm{P}\mathrm{Q}^{T} \quad s.t, \mathrm{P}\geq 0 \quad \textrm{and} \quad \mathrm{Q}\geq 0.
\end{equation}

\subsection{NMF for Traffic Pattern Elicitation}
In traffic scenario, location-time matrices $\boldsymbol{\mathrm{X}} \in {\mathbb{R}} ^{(N \times M)}$ (which stores the 2019 traffic record) and $\boldsymbol{\mathrm{Y}} \in {\mathbb{R}} ^{(A \times B)}$ (which stores the 2020 traffic record) are the representation of the count of vehicles with respect to given location and time. $N$, $A$ in  is the total number of unique locations i.e. $12656$ locations in 2019, $6537$ locations in 2020; and $M$, $B$ is the range of time records i.e. $12$ hours from 07:00 to 18:00 hours with the interval of 1 hour. 

NMF is very useful in separately extracting the spatio-temporal patterns by representing (location $\times$ features) and (time $\times$ features) individually for 2019 and 2020, in which each column of the factor matrices is a pattern or feature. Based on the rank $r$ defined in Eq \ref{equation1} and $k$ in Eq \ref{equation2}, $r$ and $k$ distinctive patterns can be obtained from $\boldsymbol{\mathrm{X}}$ and $\boldsymbol{\mathrm{Y}}$ during the factorization process respectively.

\par From Eq \ref{equation3} \& \ref{equation4}, the optimization problem is formulated as a minimization problem with Euclidean distance as the cost function individually for 2019 and 2020 as below,

\begin{equation}
\label{equation3}
\min_{\mathrm{U \geq 0,V \geq 0}}  f(\mathrm{U},\mathrm{V}) = \left \|\boldsymbol{\mathrm{X}} - \mathrm{U}\mathrm{V}^{T}  \right \|_{2} \quad s.t, \mathrm{U}\geq 0 \quad \textrm{and} \quad \mathrm{V}\geq 0
\end{equation}

\begin{equation}
\label{equation4}
\min_{\mathrm{P \geq 0,Q \geq 0}}  f(\mathrm{P},\mathrm{Q}) = \left \|\boldsymbol{\mathrm{Y}} - \mathrm{P}\mathrm{Q}^{T}  \right \|_{2} \quad s.t, \mathrm{P}\geq 0 \quad \textrm{and} \quad \mathrm{Q}\geq 0
\end{equation}

Determination of ranks to run the NMF model on both of the generated location-time matrices is discussed elaborately in section \ref{evaluationmeasures}. Nonnegative lower-dimensional representation of the matrices $\boldsymbol{\mathrm{X}}$ and $\boldsymbol{\mathrm{Y}}$ 
generated on performing the NMF model contains (location $\times$ patterns) and (time  $\times$ patterns) as $\mathrm{U}$ and $\mathrm{V}$ for $\boldsymbol{\mathrm{X}}$, $\mathrm{P}$ and $\mathrm{Q}$ for $\boldsymbol{\mathrm{Y}}$. These NMF-generated location and time patterns are grouped into individual location clusters and time clusters, respectively.    

\subsection{Understanding Traffic Pattern Variations}
Spatio-temporal traffic patterns in ITS are much essential for better understanding the latent spatio-temporal variation patterns. These observations of pattern variations help us to predict or manage the traffic scenario prior to various application based use-cases and help to improve the traffic management across the country which further leads to the development of smart environment in ITS. The temporal variation patterns observed in 2019 are compared with the corresponding spatial variation patterns as well as compared to the temporal variation patterns of 2020. This pattern comparison is very useful in unexpected scenarios such as COVID-19 pandemic, and helps to get various insights of the vehicular traffic pattern distributions.

\section{A Case Study}
\label{section4}
\subsection{Dataset}
\begin{figure}
     \centering
     \begin{subfigure}[b]{0.475\textwidth}
         \centering
         \includegraphics[width=\textwidth]{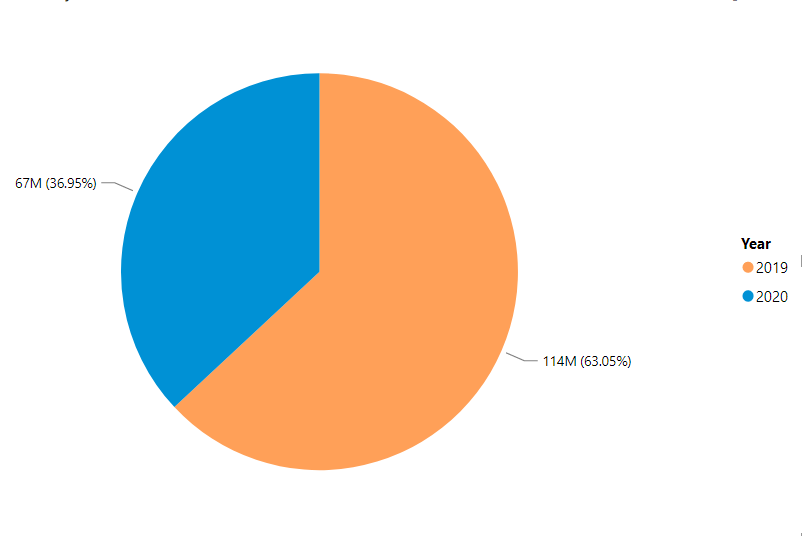}
         \caption{Spread of count of vehicles in Great Britain during 2019 and 2020}
         \label{fig:2a}
     \end{subfigure}
     \hfill
     \begin{subfigure}[b]{0.475\textwidth}
         \centering
         \includegraphics[width=\textwidth]{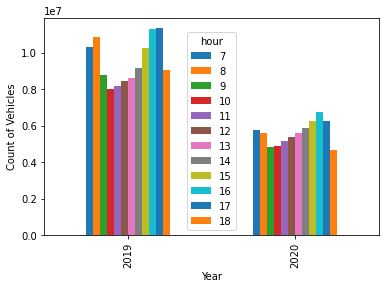}
         \caption{Hourly comparison of vehicle counts in the year 2019 and 2020}
         \label{fig:2b}
     \end{subfigure}
     \caption{Spread of vehicle counts in Great Britain}
     \label{fig:2}
\end{figure}

The case study is performed on the dataset \cite{ds}, which is downloaded from the repository of the Department of Transport, Great Britain. Dataset contains the count of vehicles that is recorded in major and minor roads of Great Britain. The data is filtered to include traffic records for the year 2019 and 2020 only. Fig. \ref{fig:2} shows the traffic spread of the count of various kinds of passing vehicles that were recorded from $0700$ to $1800$ hours during the year 2019 and 2020 across Great Britain. This figure explicitly shows that there is a drastic decrease in the total number of vehicles on roads. However, this could not answer how patterns are varied during COVID-19.

\begin{comment}
\begin{figure}
    \centering
    \includegraphics[width = 4in, height = 2in]{figs/initial0.png}
    \caption{Average count of all motor vehicles in Great Britain during 2018}
    \label{fig:2}
\end{figure}
\end{comment}

\par To generate location-time matrices $\boldsymbol{\mathrm{X}}$ and $\boldsymbol{\mathrm{Y}}$ corresponding to the years 2019 and 2020, respectively, 12656 unique locations with 12 hours time from 2019 and 6537 unique locations with 12 hours time from 2020 are grouped according to the location \& hour and the cumulative sum of the total number of recorded motor vehicles at the given location and time is calculated such that the input matrices $\boldsymbol{\mathrm{X}}$ and $\boldsymbol{\mathrm{Y}}$ are in the order of $(12656 \times 12)$ and ($(6537 \times 12)$), respectively.

The generated matrices contain numerical values, which need to be normalized in prior to the NMF modeling to make NMF perform better on the generated location-time matrices. The process of performing Normalization on the input matrices scales the numerical variables into nonnegative floating-point variables of range $0$ to $1$. Data Normalization is achieved by using MinMax Scalar Transforms. The normalized matrix is further analyzed to determine the rank using various evaluation measures such as  Calinski and Harabasz score which is the ratio between with-in cluster\footnote{cluster is nothing but pattern.} and between-cluster dispersion. 

\subsection{Evaluation Measures}
In follow-up towards the step of Normalization, evaluation measures are performed mainly to determine the rank that needs to be given as an input to the NMF model to generate temporal and spatial patterns for 2019 and 2020. To determine the optimal rank, with-in cluster dispersion (Eq \ref{equation5}) and between-cluster dispersion (Eq \ref{equation6}) analysis is carried out on the clusters and datapoints. The value should be low for with-in cluster dispersion evaluation and high for between-cluster dispersion evaluation which in-turns proves that the clusters should be well separated from other clusters and the datapoints within cluster should be grouped together, respectively. 

\begin{figure}[ht!]
     \centering
     \begin{subfigure}[t]{0.475\textwidth}
         \centering
         \includegraphics[width=\textwidth]{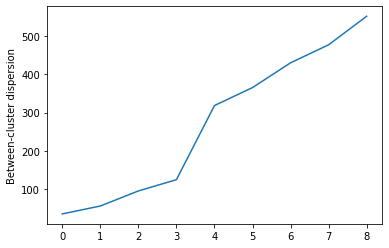}
         \caption{Between-cluster dispersion 2019}
         \label{fig:bcd2019}
     \end{subfigure}
     \hfill
     \begin{subfigure}[t]{0.475\textwidth}
         \centering
         \includegraphics[width=\textwidth]{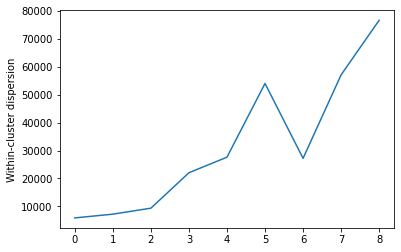}
         \caption{With-in cluster dispersion 2019}
         \label{fig:wcd2019}
     \end{subfigure}
     \caption{Rank evaluation measures for 2019. (x-axis indicates number of clusters/patterns.}
     \label{fig:bcdwcd2019}
\end{figure}

\begin{equation}
\label{equation5}
W_{d} = \sum_{g=1}^{m}\sum_{x\epsilon C_{g}}^{}(x-c_{g})(x-c_{g})^{T}
\end{equation}

\begin{equation}
\label{equation6}
B_{d} = \sum_{g=1}^{m}n_{g}(c_{g}-c_{S})(c_{g}-c_{S})^{T}
\end{equation}
where $B_{d}$ is between-cluster dispersion, $W_{d}$ is within-cluster dispersion, $S$ indicates the set of data, $g$ is clusters, $C_{g}$ is the set of points in the cluster, $c_{g}$ is the center of the cluster, $c_{S}$ is the center of data, and $n_{g}$ is the number of points in the cluster. 

Fig. \ref{fig:bcdwcd2019} shows the with-in cluster dispersion and between-cluster dispersion evaluation for the year 2019. From the figure \ref{fig:bcd2019}, it is clear that there is an increase in the trend of between-cluster dispersion graph values $3$, $4$ and $6$. In addition to this, from fig. \ref{fig:wcd2019}, the corresponding values $3$, $4$ and $6$ are analysed and it is noticeable that within-cluster value of $6$ is quite convincing as the value trend should be opposite to that of the between-cluster value trend. In this scenario, there is a hike in the value $6$ of fig. \ref{fig:bcd2019} and corresponding drop in the value $6$ of fig. \ref{fig:wcd2019}. From this analysis, one can conclude that the association of data points is good at the value $6$, hence the rank value is set to be as $6$ for running the NMF model in 2019.

\begin{figure}[ht!]
     \centering
     \begin{subfigure}[ht!]{0.475\textwidth}
         \centering
         \includegraphics[width=\textwidth]{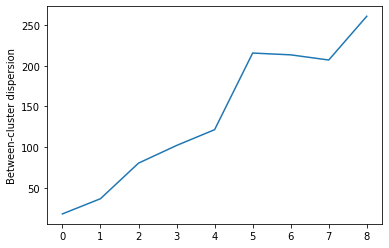}
         \caption{Between-cluster dispersion 2020}
         \label{fig:bcd2020}
     \end{subfigure}
     \hfill
     \begin{subfigure}[ht!]{0.475\textwidth}
         \centering
         \includegraphics[width=\textwidth]{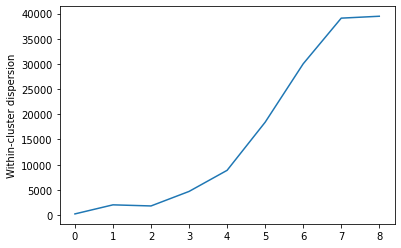}
         \caption{With-in cluster dispersion 2020}
         \label{fig:wcd2020}
     \end{subfigure}
     \caption{Rank evaluation measures for 2020}
     \label{fig:bcdwcd2020}
\end{figure}
\label{evaluationmeasures}
For the 2020 datapoints, with-in cluster dispersion and between-cluster dispersion are carried out and the evaluation measures are recorded in fig. \ref{fig:bcdwcd2020}. The between-cluster dispersion values are varying across the values $2$, $4$, $5$ and $7$ and corresponding with-in cluster dispersion values are to be taken into consideration in determining the rank to perform NMF model on the 2020 traffic datapoints. From the evaluation, it is obvious to take the rank value as $4$ to compare the relationship of the spatio-temporal patterns. This already hints that in 2020 during COVID-19, two spatio-temporal patterns have disappeared. Further analysis is needed to find out what that pattern is?

\subsection{Traffic Pattern Results for 2019}
\begin{figure}
    \centering
    \includegraphics[width = 4in, height = 2in]{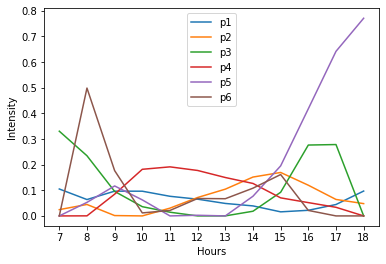}
    \caption{Temporal Patterns in 2019}
    \label{fig:ti2019}
\end{figure}

Fig. \ref{fig:ti2019} shows the traffic patterns obtained from the lower-dimensional factor matrix $\mathrm{V}$ on performing NMF model in normalized $\boldsymbol{\mathrm{X}}$ matrix belonging to 2019 traffic records. According to the rank obtained from evaluation measures, there are 6 temporal patterns (p1 to p6) observed. The x-axis in Fig. \ref{fig:ti2019} indicates record from $07:00$ to $18:00$ hours, and y-axis indicates the intensity of vehicles at the given time. On analysing the observed patterns, it is very clear that pattern (p6) records high traffic data in the early morning, pattern (p5) records high traffic data in the evening followed by pattern (p3) which records the second highest traffic count in the evening. The pattern (p2) shows the decrease of traffic density and moderate increase of the traffic density in morning and evening, respectively. Comparatively, the pattern (p1) shows very low traffic density record during evening peak times. In contrast to all, there is an abnormal traffic behaviour pattern in (p4) during the period $10:00$ to $14:00$ hours in comparison to all other pattern behaviours.   

\begin{comment}
\begin{figure}
     \centering
     \begin{subfigure}[b]{0.475\textwidth}
         \centering
         \includegraphics[width=\textwidth]{figs/timepatterns2019.png}
         \caption{time patterns 2019}
         \label{fig:ti2019}
     \end{subfigure}
     \hfill
     \begin{subfigure}[b]{0.475\textwidth}
         \centering
         \includegraphics[width=\textwidth]{figs/timepatterns2020with6clusters.png}
         \caption{time patterns 2020 with 6 clusters}
         \label{fig:lo2019}
     \end{subfigure}
     \caption{patterns 2019}
     \label{fig:ti2019}
\end{figure}
\end{comment}

\begin{figure}[ht!]
    \centering
    \includegraphics[width=.27\textwidth]{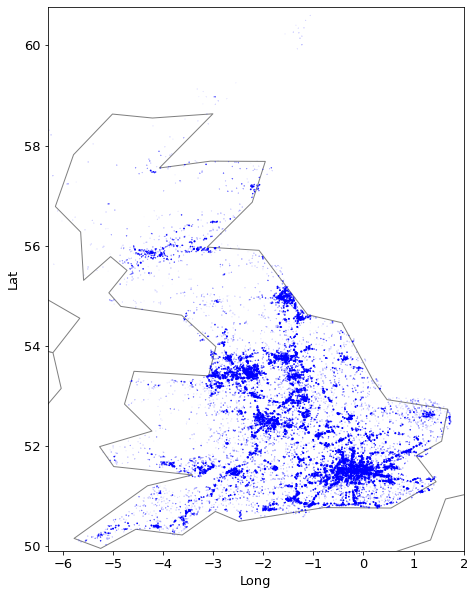}\hfill
    \includegraphics[width=.27\textwidth]{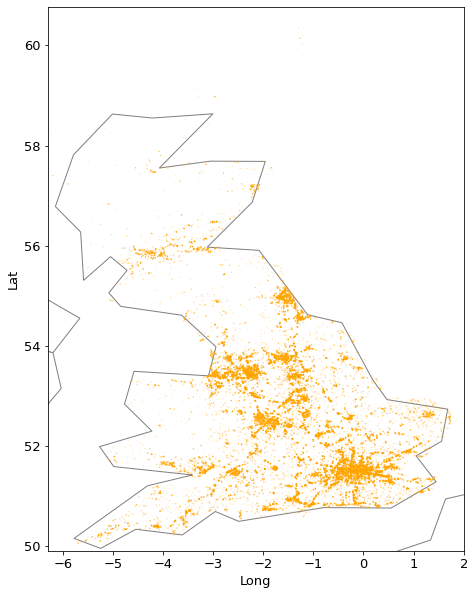}\hfill
    \includegraphics[width=.27\textwidth]{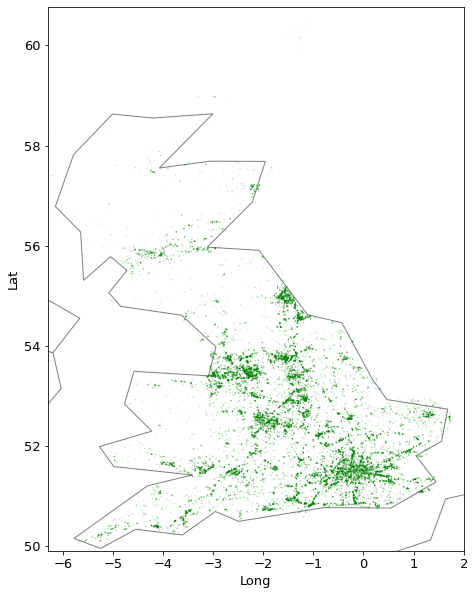}\hfill
    \\[\smallskipamount]
    \includegraphics[width=.27\textwidth]{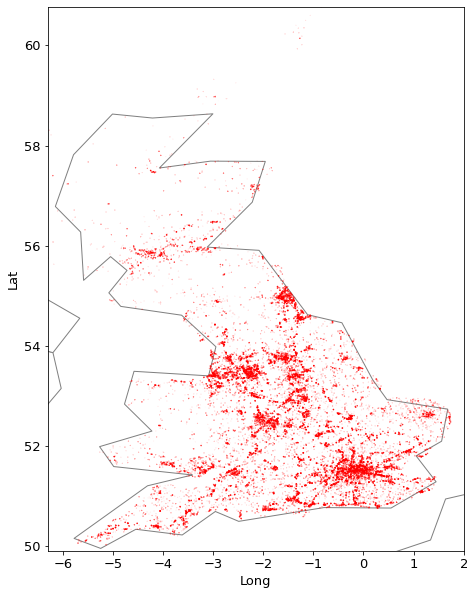}\hfill
    \includegraphics[width=.27\textwidth]{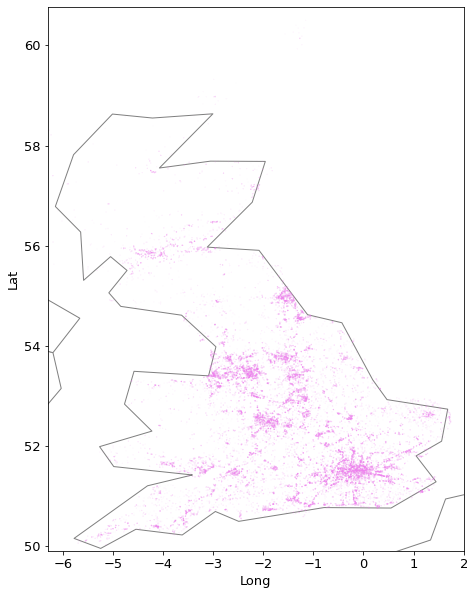}\hfill
    \includegraphics[width=.27\textwidth]{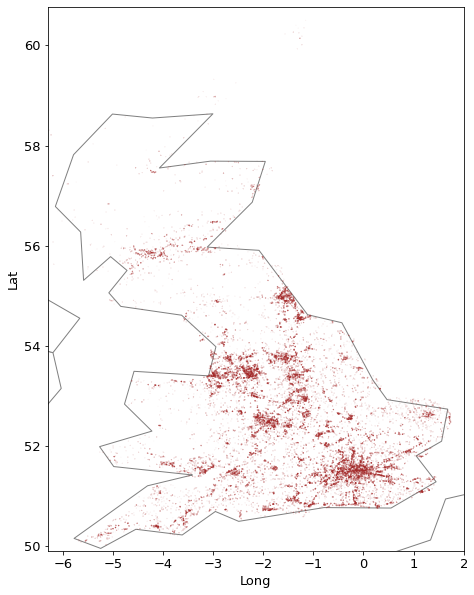}\hfill
    \\[\smallskipamount]
    \caption{Spatial patterns in 2019 with 6 clusters}\label{fig:lo2019}
\end{figure}

These pattern visualizations show the different temporal behaviours of people's mobility. The insights from fig. \ref{fig:ti2019} are: the number of people who are active in the morning are not very much active in the evening, and vice-versa. This is justified with the insights from p6 and p2, respectively.   
The corresponding spatial behavioural patterns in 2019 are obtained from the lower-dimensional location factor matrix $\mathrm{U}$ on performing NMF model in the normalized $\boldsymbol{\mathrm{X}}$ matrix in relation to the above temporal patterns. In terms of spatial relationship, there is only a minor difference between the spatial patterns as observed in fig. \ref{fig:lo2019}. However, if we look into spatial pattern (purple colour) in comparison with the temporal pattern (p5) from fig. \ref{fig:ti2019}, there is a difference such that the pattern (p5) shows the peak in the evening but in parallel the spread of spatial vehicle density in corresponding to the pattern (p5) is less in comparison to that of the other patterns from fig. \ref{fig:lo2019}. 

\subsection{Traffic Pattern Results for 2020}

\begin{figure}
    \centering
    \includegraphics[width = 4in, height = 2in]{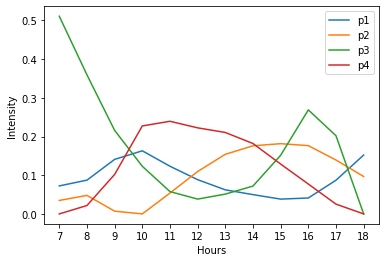}
    \caption{Temporal Variation Patterns in 2020}
    \label{fig:ti2020}
\end{figure}

\begin{comment}
\begin{figure}
     \centering
     \begin{subfigure}[t]{0.475\textwidth}
         \centering
         \includegraphics[width=\textwidth]{figs/timepatterns2020.png}
         \caption{Temporal patterns in 2020 with 4 clusters}
         \label{fig:ti2020a}
     \end{subfigure}
     \hfill
     \begin{subfigure}[t]{0.475\textwidth}
         \centering
         \includegraphics[width=\textwidth]{figs/timepatterns2020with6clusters.png}
         \caption{Temporal patterns in 2020 with 6 clusters}
         \label{fig:ti2020b}
     \end{subfigure}
     \caption{Temporal pattern observations with multi-cluster analysis}
     \label{fig:ti2020}
\end{figure}
\end{comment}

\begin{figure}[ht!]
    \centering
    \includegraphics[width=.27\textwidth]{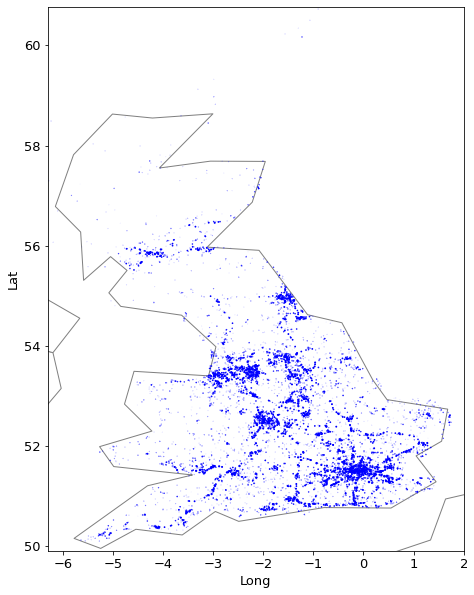}
    \includegraphics[width=.27\textwidth]{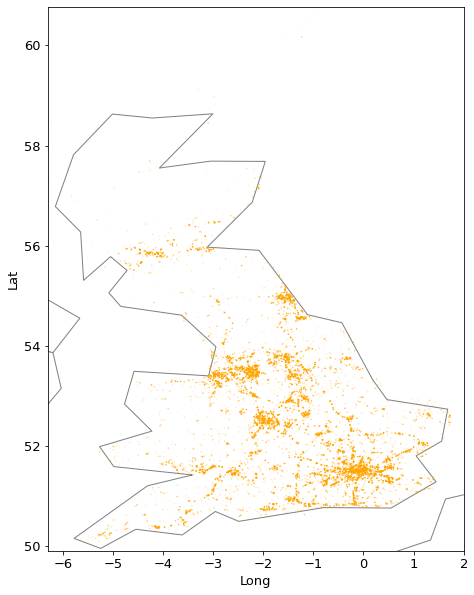}
    \includegraphics[width=.27\textwidth]{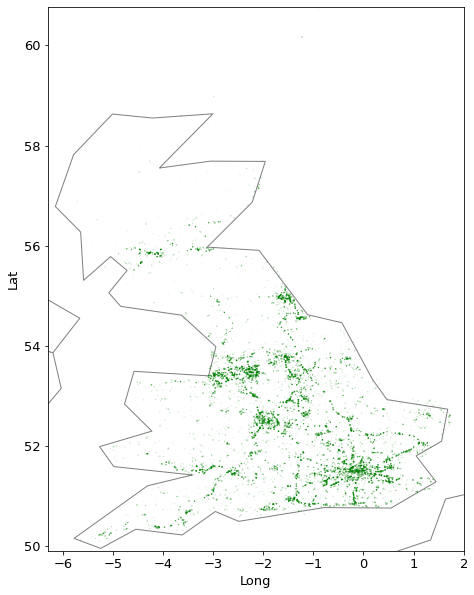}
    \\[\smallskipamount]
    \includegraphics[width=.27\textwidth]{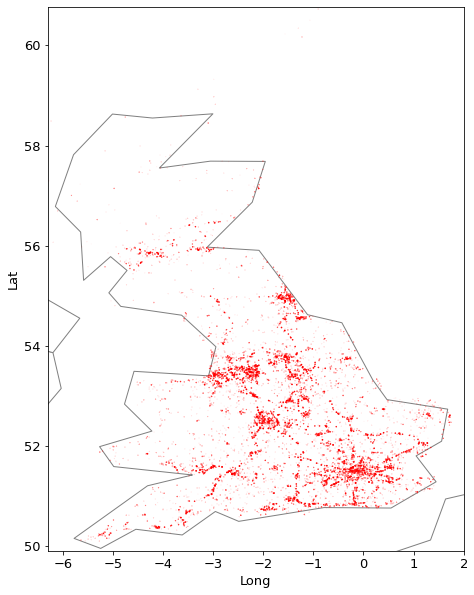}
    \\[\smallskipamount]
    \caption{Spatial patterns in 2020 with 4 clusters}\label{fig:lo2020}
\end{figure}

\begin{comment}
\begin{figure}[ht!]
    \centering
    \includegraphics[width=.27\textwidth]{figs/p12020.png}\hfill
    \includegraphics[width=.27\textwidth]{figs/p22020.png}\hfill
    \includegraphics[width=.27\textwidth]{figs/p32020.png}\hfill
    \\[\smallskipamount]
    \includegraphics[width=.27\textwidth]{figs/p42020.png}\hfill
    \includegraphics[width=.27\textwidth]{figs/p52020.png}\hfill
    \includegraphics[width=.27\textwidth]{figs/p62020.png}\hfill
    \\[\smallskipamount]
    \caption{Spatial patterns in 2020 with 6 clusters}\label{fig:lo2020}
\end{figure}
\end{comment}
Temporal pattern observations for 2020 with 4 clusters is plotted in fig. \ref{fig:ti2020}, which shows the temporal pattern elicited from the lower-dimensional representation matrix $\mathrm{Q}$ of the $\boldsymbol{\mathrm{Y}}$ matrix by running the NMF model with rank $4$. Fig. \ref{fig:ti2020} shows the temporal patterns observed during $07:00$ to $18:00$ hours during the COVID-19 pandemic year 2020, and it is very obvious that two patterns are already missing in 2020 in-comparison to the temporal patterns observed in 2019. 

Fig. \ref{fig:ti2020} shows the temporal patterns observed with 4 clusters in which the pattern (p3) records high traffic density in the morning with a gradual decrease in the afternoon period and sudden increase of traffic density in the evening peak hours. However, the overall traffic density is $52\%$ less than the year 2019. Though the other patterns (p1, p2, p4) have a gradual increase and decrease during the overall day, the pattern (p3) is dominating in the morning and evening, respectively. 

The corresponding spatial behavioural patterns in 2020, shown in fig. \ref{fig:lo2020}, obtained from the lower-dimensional location factor matrix $\mathrm{P}$ on performing NMF model in the normalized $\boldsymbol{\mathrm{Y}}$ matrix are shown in relation to the temporal patterns in fig. \ref{fig:ti2020}. According to the spatial pattern observations, though the pattern (blue) is diversely spread across the country in comparison to the other patterns, the temporal behaviours of the corresponding pattern (p1) in fig. \ref{fig:ti2020} are moderate in comparison to the other temporal pattern behaviours.

\subsection{Traffic Pattern Variation Analysis}
The importance of understanding the traffic pattern variations is much essential for building a smarter and efficient transportation system. On comparing the temporal patterns of before COVID-19 and during COVID-19 scenarios, it is very obvious that the vehicular mobility patterns of the people who used to travel in the early morning are drastically/moderately changed in the corresponding patterns in 2020, and there is a disappearance of two patterns when compared to 2019 temporal behaviours. In particular, vehicular traffic pattern (p6) is higher in the morning and slight increase in the evening with a fluctuation in-between. However, the corresponding pattern is completely not present in 2020 due to the sudden influence of COVID-19 cases. 

\par On the other hand, the temporal patterns (p3 and p4) in 2019 and 2020 show variations in behaviour accordingly. These temporal variations in certain patterns show that the people who are active in the morning were reduced and vice-versa, and to be noticed that the people's activity is increased in couple of the patterns of 2020. However, in comparison, the pattern (p1) in 2019 is almost similar to the pattern (p1) in 2020. In parallel, comparison to the spatial patterns observed in 2019 and 2020, there is a very small difference in spatial pattern variations however there is a spatio-temporal pattern variation observed within the years 2019 and 2020. The effect of pandemic plays a major role in the temporal patterns of Great Britain with only few spatial pattern density differences observed for 2019 and 2020.

\begin{comment}
\begin{figure}
    \centering
    \includegraphics[width=.27\textwidth]{figs/l120204c.png}
    \includegraphics[width=.27\textwidth]{figs/l220204c.png}\hfill
    \\[\smallskipamount]
    \includegraphics[width=.27\textwidth]{figs/l320204c.png}
    \includegraphics[width=.27\textwidth]{figs/l420204c.png}\hfill
    \\[\smallskipamount]
    \caption{Location patterns in 2020 with 4 clusters}\label{fig:7}
\end{figure}
\end{comment}

\section{Conclusion}
This paper analyses the spatio-temporal traffic variation patterns recorded between $0700$ to $1800$ hours during the year 2019 (before-COVID-19) and 2020 (during-COVID-19) in Great Britain. We also demonstrate the capability of utilizing NMF in getting various useful insights from the traffic patterns elicited from the NMF outputs. The case study conducted in this paper provides detailed traffic patterns and shows the variations in the spatio-temporal patterns. Our observations from the temporal and spatial patterns show that the temporal traffic patterns are drastically varied during COVID-19, while there is only a few minor variations of spatial traffic patterns. This spatio-temporal traffic pattern variation insight will lead to more understanding about the abnormal behaviour of the traffic mobility patterns compared to the past historical data and further leads to improvising the traffic management planning in the future pandemic or unavoidable scenarios. In future, a more detailed analysis will be conducted to see how patterns change over the period of month or week.
\label{section5}

\section*{Acknowledgement}
This research was supported by the National Research Foundation of Korea (Grant No. 2020R1A2C1012196), and in part by the School of Computer Science and Engineering, Ministry of Education, Kyungpook National University, South Korea, through the BK21 Four Project, AI-Driven Convergence Software Education Research Program, under Grant 4199990214394.

\end{document}